\documentclass[acmtog, nonacm]{acmart}
\AtBeginDocument{%
  }

\authorsaddresses{Authors' Contact Information: Yuchen Guo, Northwestern University, Evanston, USA; Junli Gong, Northeastern University, Seattle, USA; Hongmin Cai, South China University of Technology, Guangzhou, China; Yiu-ming Cheung, Hong Kong Baptist University, Hong Kong SAR, China; Weifeng Su, Beijing Normal - Hong Kong Baptist University, Zhuhai, China. Corresponding to Yuchen Guo (yuchenguo2027@u.northwestern.edu) and Weifeng Su(wfsu@bnbu.edu.cn).}

\begin{document}

\title{LumiVideo: An Intelligent Agentic System for Video Color Grading}


\author{Yuchen Guo}
\affiliation{%
  \institution{Northwestern University}
  \city{Evanston}
  \country{USA}}
\email{yuchenguo2027@u.northwestern.edu}

\author{Junli Gong}
\affiliation{%
  \institution{Northeastern University}
  \city{Seattle}
  \country{USA}}

\author{Hongmin Cai}
\affiliation{%
  \institution{South China University of Technology}
  \city{Guangzhou}
  \state{Guangdong}
  \country{China}}

\author{Yiu-ming Cheung}
\affiliation{%
  \institution{Hong Kong Baptist University}
  \city{Hong Kong SAR}
  \country{China}}

\author{Weifeng Su}
\affiliation{%
  \institution{Beijing Normal - Hong Kong Baptist University}
  \city{Zhuhai}
  \state{Guangdong}
  \country{China}}

\renewcommand{\shortauthors}{Guo et al.}

\begin{abstract}
Video color grading is a critical post-production process that transforms flat, log-encoded raw footage into emotionally resonant cinematic visuals. Existing automated methods act as static, black-box executors that directly output edited pixels, lacking both interpretability and the iterative control required by professionals. We introduce \textbf{\textit{LumiVideo}}, an agentic system that mimics the cognitive workflow of professional colorists through four stages: Perception, Reasoning, Execution, and Reflection. Given only raw log video, LumiVideo autonomously produces a cinematic base grade by analyzing the scene's physical lighting and semantic content. Its Reasoning engine synergizes an LLM's internalized cinematic knowledge with a Retrieval-Augmented Generation (RAG) framework via a Tree of Thoughts (ToT) search to navigate the non-linear color parameter space. Rather than generating pixels, the system compiles the deduced parameters into industry-standard ASC-CDL configurations and a globally consistent 3D LUT, analytically guaranteeing temporal consistency. An optional Reflection loop then allows creators to refine the result via natural language feedback. We further introduce \textit{\textbf{LumiGrade}}, the first log-encoded video benchmark for evaluating automated grading. Experiments show that LumiVideo approaches human expert quality in fully automatic mode while enabling precise iterative control when directed.
\vspace{\baselineskip}

\noindent \textbf{Project page:} \url{https://eurekaarrow.github.io/LumiVideo/}
\end{abstract}



\begin{teaserfigure}
  \includegraphics[width=\textwidth]{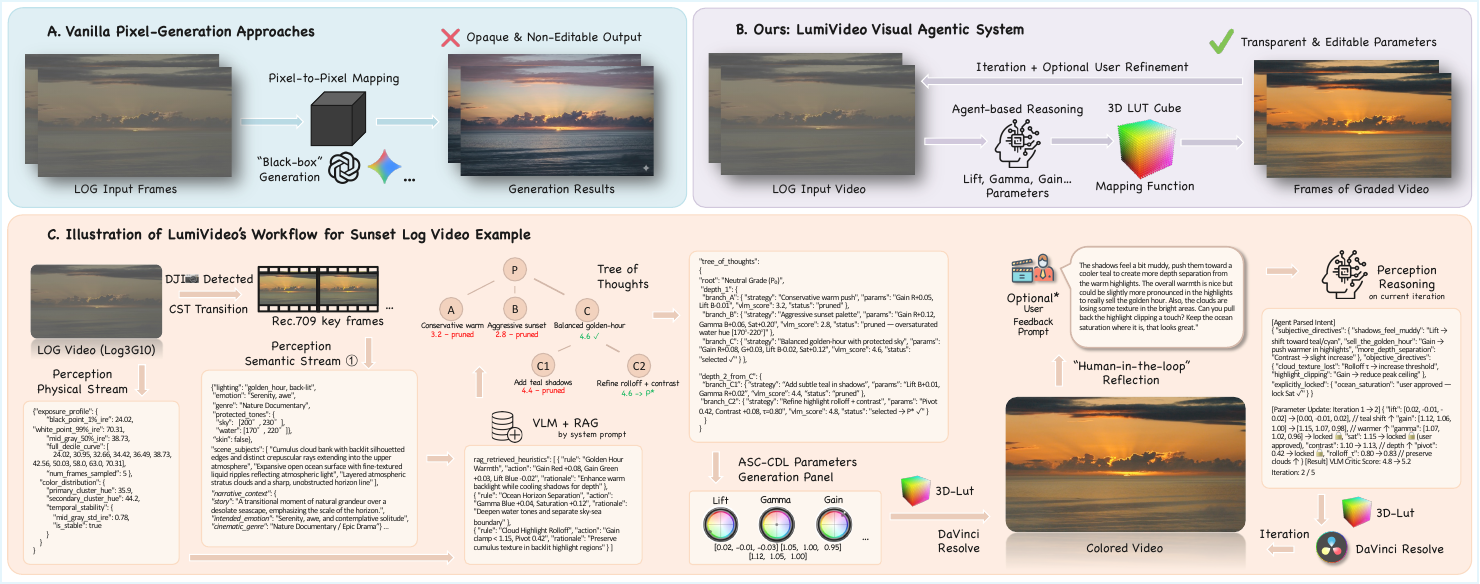}
  \caption{From Pixel Generation to Parameter Reasoning. \textcircled{1}: The perception semantic stream JSON is part of an example.}
  \label{fig:teaser}
\end{teaserfigure}

\maketitle

\section{Introduction}
Video color grading is the cornerstone of cinematic post-production \cite{bonneel2013example, shin2025video, dancyger2018technique}. It is a highly intricate process that transforms flat, log-encoded raw footage into visually compelling and emotionally resonant storytelling mediums. Unlike basic color correction, which merely neutralizes exposure and white balance, professional color grading requires a deep understanding of lighting physics, color harmony, and narrative intent. Traditionally, this process relies on the specialized expertise of human colorists operating complex, non-linear editing systems (e.g., DaVinci Resolve) to manipulate high-dynamic-range (HDR) color spaces. Given the sheer volume of video content generated today, there is an urgent demand for intelligent systems capable of automating this labor-intensive workflow without compromising artistic fidelity.

Recently, the rapid evolution of Vision-Language Models (VLMs) and multi-modal foundation models has sparked immense interest in automated visual editing \cite{zhang2023adding}. Existing methodologies predominantly formulate visual stylization as an image-to-image translation task \cite{brooks2023instructpix2pix} or rely on generative diffusion models \cite{you2025photoframer} to directly synthesize the edited frames. However, when applied to professional video color grading, this ``pixel-generation'' paradigm reveals fatal flaws. First, end-to-end models operate as opaque black boxes, completely destroying the mathematical interpretability of the color space \cite{chen2025photoartagent, yao2026photoagent}. Second, generating pixels frame-by-frame inevitably introduces severe temporal flickering and spatial artifacts, rendering the output unusable for video. Most critically, these static executors are entirely disconnected from industry-standard workflows; they output rigid images rather than manipulable color parameters, thereby denying creators the essential ability to perform precise, state-aware iterative refinements.

To bridge the chasm between generative AI and the professional cinematic pipeline, we argue that intelligent color grading must be reformulated from a pixel-generation task into a transparent, parameter-driven reasoning process. Inspired by the cognitive workflow of human colorists---who first \textit{read} the footage, then \textit{reason} about the target aesthetic, \textit{execute} adjustments on the color wheels, and \textit{refine} based on director feedback---we introduce \textbf{\textit{LumiVideo}}, a novel agentic system that decomposes the color grading lifecycle into four interactive stages: Perception, Reasoning, Execution, and Reflection. By shifting the model's output from arbitrary RGB arrays to industry-standard grading parameters, LumiVideo guarantees absolute temporal consistency and zero structural degradation.

A key design principle of LumiVideo is \textbf{autonomy}: the system produces a complete, cinematic base grade from raw log footage without requiring any user input before iteration. During the Perception stage, a dual-stream module leverages VLMs alongside deterministic physical algorithms (e.g., Color Space Transforms) to decouple the high-level semantic context---including scene lighting, time-of-day cues, and protected subjects such as skin tones---from the raw image statistics. The core of our system lies in the Reasoning stage, where a Large Language Model (LLM) acts as the Look Developer Agent. Given only the perceived scene state, the LLM autonomously infers the appropriate cinematic treatment by synergizing its internalized visual knowledge with a professional heuristic Retrieval-Augmented Generation (RAG) framework. Recognizing the vast and highly non-linear nature of color parameter spaces, we implement a Tree of Thoughts (ToT) strategy, enabling the LLM to systematically explore, evaluate, and deduce the optimal color decisions. For Execution, the agent compiles these decisions into mathematically rigorous ASC-CDL (Color Decision List) configurations and a globally consistent 3D Look-Up Table (LUT). Finally, the Reflection stage introduces an optional state-aware loop: when a director wishes to refine the base grade (e.g., \textit{``push the shadows toward teal''}), the system updates only the relevant mathematical parameters while locking all others, enabling precise, incremental control without regenerating the underlying video.

In summary, the main contributions of this work are:

\begin{itemize}
    \item \textbf{Autonomous Cinematic Agent.} We propose LumiVideo, the first agentic system capable of producing a professional-quality cinematic base grade from raw log-encoded video without any user input, by reformulating color grading as a four-stage visual agent workflow (Perception, Reasoning, Execution, Reflection).
    \item \textbf{Structured Cinematic Reasoning.} We introduce a novel Reasoning module that synergizes an LLM with a domain-specific RAG framework via a Tree of Thoughts (ToT) search strategy, enabling systematic exploration of the highly non-linear ASC-CDL parameter space to deduce optimal grading configurations.
    \item \textbf{Industry-Standard Output with Iterative Control.} LumiVideo outputs strictly compliant ASC-CDL parameters compiled into globally consistent 3D LUTs, analytically guaranteeing temporal consistency and seamless integration with professional NLEs. An optional natural language-driven Reflection loop enables state-aware iterative refinement with precise, predictable control.
    \item \textbf{LumiGrade Benchmark.} To support research in this underexplored domain, we introduce LumiGrade, a benchmark of over 100 professionally captured, log-encoded video clips spanning diverse cameras and scenes with rich metadata annotations---to our knowledge, the first publicly available benchmark for evaluating automated video color grading from raw footage.
\end{itemize}

\section{Related Work}
\label{sec:related}

\subsection{Color Grading and Color Transfer}
\label{subsec:rw_color}
Classical color transfer methods match statistical distributions between images via mean-variance alignment~\cite{reinhard2002color} or optimal transport~\cite{pitie2005n, pitie2007automated}, but operate globally without iterative control. Deep learning introduced learnable 3D LUTs~\cite{zeng2020learning} with extensions to compressed~\cite{zhang2022clut} and spatially-aware~\cite{wang2021real, yang2022adaint} representations. For video, Bonneel et al.~\cite{bonneel2013example} addressed temporal coherence, and Shin et al.~\cite{shin2025video} proposed diffusion-based LUT generation from reference images. However, all these methods are \textit{single-shot executors}: they produce no interpretable, editable parameters and cannot support targeted iterative adjustments. None output industry-standard formats (e.g., ASC-CDL) for integration with professional NLEs. Moreover, existing datasets~\cite{bain2020condensed, shin2025video} consist of post-produced sRGB clips, bypassing the core challenge of grading from raw log-encoded footage. Our work departs by producing standard-compliant parameters with state-aware iterative refinement, accompanied by a dedicated log-encoded benchmark.

\begin{figure*}[t]
    \centering
    \includegraphics[width=\textwidth]{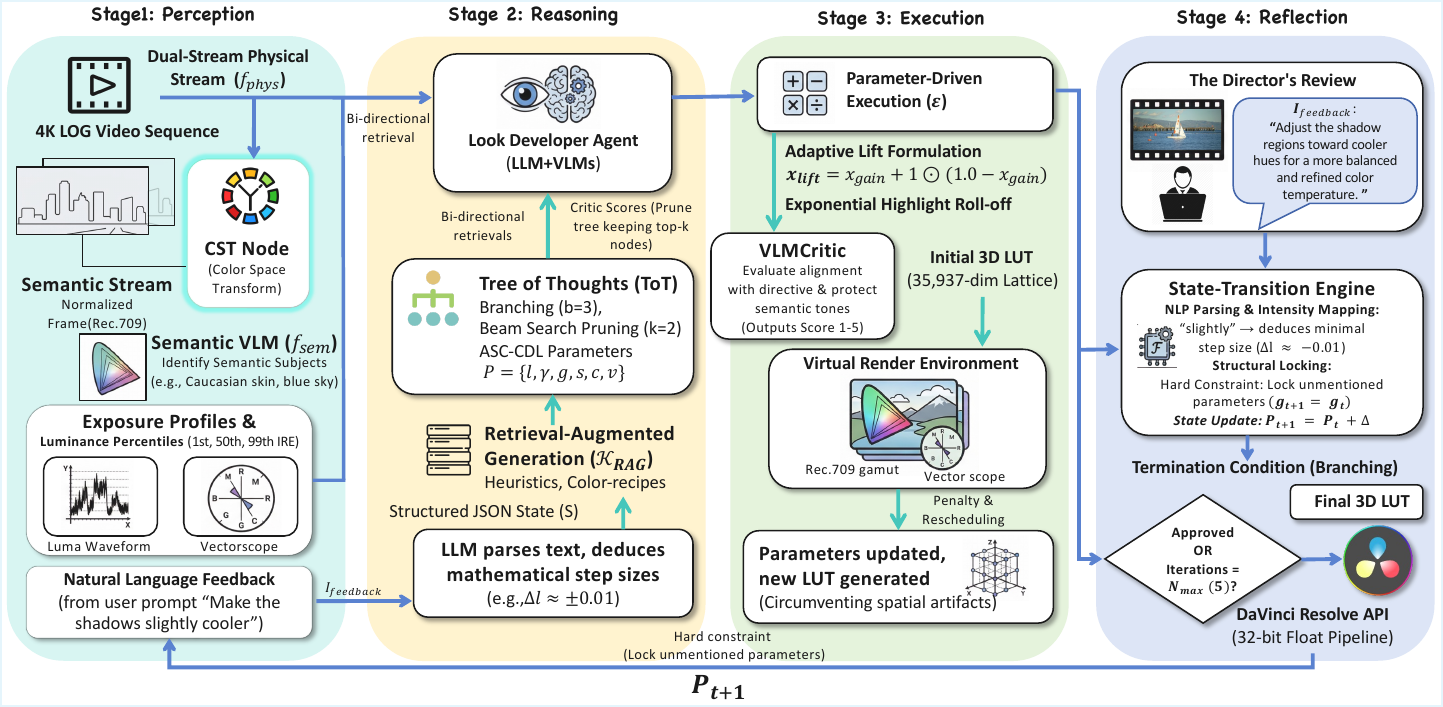}
    \caption{Architecture of LumiVideo. As an example, we set up max iteration times to 5 here.}
    \label{fig:architecture}
\end{figure*}

\subsection{LLM/VLM-Based Agents for Visual Editing}
\label{subsec:rw_agent}
LLMs and VLMs~\cite{achiam2023gpt, bai2025qwen3} have enabled agentic approaches to visual editing \cite{li2025sam3, guo2025dae, wang2025multi}. Pixel-generation methods such as InstructPix2Pix~\cite{brooks2023instructpix2pix} and diffusion-based editors~\cite{zhang2023adding, you2025photoframer} lack interpretability and suffer from temporal inconsistency when applied to video. Agent-based alternatives like AgenticIR ~\cite{zhu2024intelligent} and RestoreAgent \cite{chen2024restoreagent} orchestrate tool pipelines for image restoration but operate over black-box models rather than interpretable parameters. The closest work, PhotoArtAgent~\cite{chen2025photoartagent}, outputs parameters to Adobe Lightroom with a reflection loop, validating the agent paradigm for parameter-driven editing, but PhotoArtAgent addresses only single images. PhotoAgent~\cite{yao2026photoagent} concurrently explores agentic photo editing with aesthetic planning but likewise targets single images without video-level consistency.

\subsection{Reasoning Strategies for LLM Agents}
\label{subsec:rw_reasoning}
Chain-of-Thought prompting~\cite{wei2022chain} improves LLM reasoning through intermediate steps, while Tree of Thoughts~\cite{yao2023tree} enables multi-path exploration with self-evaluation. Retrieval-Augmented Generation (RAG)~\cite{lewis2020retrieval} grounds LLM outputs in external knowledge. These techniques have been validated in NLP and code generation but remain unexplored for continuous, non-linear visual parameter spaces. LumiVideo is, to our knowledge, the first system to synergize ToT-based search with domain-specific RAG for cinematic color parameter optimization.

\section{Proposed Method}
\label{sec:method}

In this section, we detail the architecture of \textit{LumiVideo}, an agentic framework designed to transform opaque pixel-generation tasks into a transparent, parameter-driven reasoning process for video color grading. We first formulate the problem and motivate our parameter space selection (Sec.~\ref{subsec:framework}). Subsequently, we elaborate on the pipeline: the dual-stream Perception module (Sec.~\ref{subsec:perception}), the core Reasoning engine utilizing a Tree of Thoughts (ToT) and Retrieval-Augmented Generation (RAG) (Sec.~\ref{subsec:reasoning}), the Execution stage for ASC-CDL mathematical compilation (Sec.~\ref{subsec:execution}), and the Reflective iteration loop (Sec.~\ref{subsec:reflection}). Figure~\ref{fig:architecture} provides an architectural overview of the complete grading lifecycle.

\subsection{Overall Framework and Parameter Space}
\label{subsec:framework}

Unlike traditional diffusion models that learn an end-to-end mapping in the highly dimensional pixel space, \textit{LumiVideo} reframes color grading as a sequential decision-making process within a constrained, industry-standard parameter space. 

Let the input log-encoded video be denoted as $V_{log} \in \mathbb{R}^{T \times H \times W \times 3}$. Our primary objective is to deduce a globally consistent 3D Look-Up Table (LUT), $\mathcal{L}$, that maps the raw color space to a cinematically plausible aesthetic. In the default \textit{automatic mode}, this target aesthetic is inferred entirely from the scene content; an optional textual directive $I_{dir}$ may be supplied to override or bias the target style toward a specific creative intent. The generation of $\mathcal{L}$ is governed by a set of color parameters $\mathbf{P}$. 

\textbf{Why ASC-CDL?} Recent works like PhotoArtAgent parameterize their action space using Adobe Lightroom sliders (e.g., Exposure, Shadows, Clarity). However, such parameterizations are highly proprietary, photograph-centric, and non-linear, making them difficult to replicate perfectly outside the Adobe ecosystem. Alternatively, HSL (Hue, Saturation, Lightness) curves often lead to color breaking and banding in video if manipulated aggressively. To circumvent these issues, we strictly adopt the ASC-CDL (American Society of Cinematographers Color Decision List) standard. ASC-CDL is mathematically transparent, globally supported by all professional NLEs (Non-Linear Editors) like DaVinci Resolve, and natively hardware-accelerated. \enlargethispage{\baselineskip} We formulate our action space as:
$$\mathbf{P} = \{ \mathbf{l}, \mathbf{\gamma}, \mathbf{g}, s, c, v \}$$
where $\mathbf{l}, \mathbf{\gamma}, \mathbf{g} \in \mathbb{R}^3$ represent the Lift, Gamma, and Gain operations for the RGB channels. $s, c, v \in \mathbb{R}$ represent saturation, contrast, and pivot. 

The workflow is orchestrated through a pipeline $\mathcal{A} = \langle \mathcal{P}, \mathcal{R}, \mathcal{E}, \mathcal{F} \rangle$. The Perception module $\mathcal{P}$ extracts physical and semantic scene descriptors; the Reasoning agent $\mathcal{R}$ queries the RAG database and structures a ToT to autonomously search for the optimal $\mathbf{P}$; the Execution module $\mathcal{E}$ compiles these into a 3D LUT; and the optional Reflection module $\mathcal{F}$ enables incremental parameter adjustments via natural language feedback without regenerating the spatial pixels.

\subsection{Dual-Stream Perception Module ($\mathcal{P}$)}
\label{subsec:perception}

A fundamental challenge in multi-modal video processing is the visual hallucination caused by log-encoded input (e.g., RED Log3G10). Log curves compress dynamic range, resulting in flat, desaturated images that VLMs frequently misinterpret as foggy or underexposed environments. The Perception module $\mathcal{P}$ acts as the agent's \textit{scene reading} stage, extracting a decoupled state $S = \{f_{phys}, f_{sem}\}$ that grounds all downstream reasoning.

\textbf{Physical Stream ($f_{phys}$):} We apply a deterministic Color Space Transform (CST) with CAT02 chromatic adaptation to map the log anchor frame to Rec.709. From this normalized space, we compute zone-based luminance percentiles (1st, 50th, 99th IRE) to define the absolute black point, mid-gray, and white point, providing the agent with an objective exposure profile independent of any VLM interpretation.

\textbf{Semantic Stream ($f_{sem}$):} Concurrently, a VLM analyzes the normalized frame to extract the high-level scene understanding that drives the agent's creative decisions. We design a structured prompt template that instructs the VLM to act as a \textit{Scene Analyst}, requiring it to output a JSON dictionary covering four categories: (1)~\textit{lighting and mood} (e.g., golden hour, overcast, artificial), (2)~\textit{narrative context} (genre, intended emotion), (3)~\textit{scene subjects} with natural language descriptions, and (4)~\textit{protected tone regions} specified as hue-angle ranges (e.g., \texttt{\{"skin": [15\textdegree, 45\textdegree], "sky": [200\textdegree, 230\textdegree]\}}). These protected regions serve as hard constraints that propagate through the entire pipeline, preventing downstream grading operations from contaminating perceptually critical hues. 

\textbf{Agent Decision Logic:} The structured state $S$ serves as the agent's \textit{world model}. In automatic mode, the Perception module further performs a scene classification step: based on the detected lighting condition and narrative context, it selects an initial retrieval strategy for the downstream RAG query (e.g., mapping ``golden\_hour + nature documentary'' to the query template ``warm cinematic grade with highlight preservation''). This decision directly shapes which heuristics the Reasoning agent retrieves, establishing a causal chain from perception to action. The complete state $S$ is serialized as JSON and injected into all downstream LLM/VLM prompts to provide persistent contextual grounding.

\subsection{Reasoning via Tree of Thoughts ($\mathcal{R}$)}
\label{subsec:reasoning}

The Reasoning stage serves as the central decision-making module. Formally, the agent seeks $\mathbf{P} = \mathcal{R}(S, \mathcal{K}_{RAG}, [I_{dir}])$, where $[\cdot]$ denotes an optional input. When no directive is provided, the agent infers the appropriate cinematic treatment entirely from $S$---for instance, detecting golden-hour backlighting in $f_{phys}$ and skin tones in $f_{sem}$ to autonomously select a warm, highlight-preserving palette. When $I_{dir}$ is supplied, it acts as a stylistic constraint that biases the search toward the specified aesthetic. Given the non-linear dynamics of color math, we formulate the parameter generation as a search problem over a Tree of Thoughts (ToT). \textit{LumiVideo} synergizes the LLM's internalized parametric knowledge (acquired during pre-training) with external non-parametric heuristics from RAG.

\textbf{Tree Structure and RAG Grounding:} Each node in the tree encapsulates a state defined by the current ASC-CDL parameters $\mathbf{P}_t$ and a natural language rationale. Before expansion, we retrieve the top-$k$ ($k=3$) heuristics from a database $\mathcal{K}_{RAG}$ (e.g., \textit{``Text: Cinematic Teal; Action: Lift Blue +0.02, Gain Red +0.05''}) using an embedding model based on cosine similarity. In automatic mode, the retrieval query is derived from the semantic descriptors in $f_{sem}$ (e.g., ``golden hour exterior, skin tones present''); when $I_{dir}$ is provided, it is used directly as the query.

\textbf{LLM Expansion Mechanism:} At each step, the LLM expands the tree with a branching factor of $b=3$. To generate mathematically grounded parameters, the expansion prompt explicitly concatenates the retrieved heuristics $\mathcal{K}_{retrieved}$, the serialized state $S$ (containing the protected tone constraints), and the parent node's rationale. The LLM is instructed to output three distinct candidate parameter sets along with explicit justifications, adapting the rigid RAG rules to the specific lighting constraints defined in $S$.

\textbf{VLM-Driven Evaluation and Pruning:} To evaluate the candidate parameters accurately, we explicitly avoid relying on the VLM's imagination. Instead, we compile each candidate parameter set $\mathbf{P}_{candidate}$ into a temporary proxy LUT and apply it to the normalized anchor frame. This rendered low-resolution preview image, alongside the constraint vector from $f_{sem}$ (and $I_{dir}$ if provided), is fed into the VLM critic. The critic outputs a quantitative score (1-5) evaluating both cinematic quality and preservation of protected tones. In automatic mode, the critic judges whether the result is cinematically plausible given the detected scene context; when $I_{dir}$ is present, it additionally evaluates alignment with the directive. To maintain high generation quality while preventing combinatorial explosion, we align our pruning strategy with a rank-based Beam Search. Nodes are sorted by their critic scores, and only the top-$k$ candidate nodes (with beam width $k=2$) are retained for the next depth of expansion. This ensures the final $\mathbf{P}^*$ maintains structural image integrity while producing a cinematically compelling grade.

\begin{figure*}[t]
    \centering
    \includegraphics[width=\textwidth]{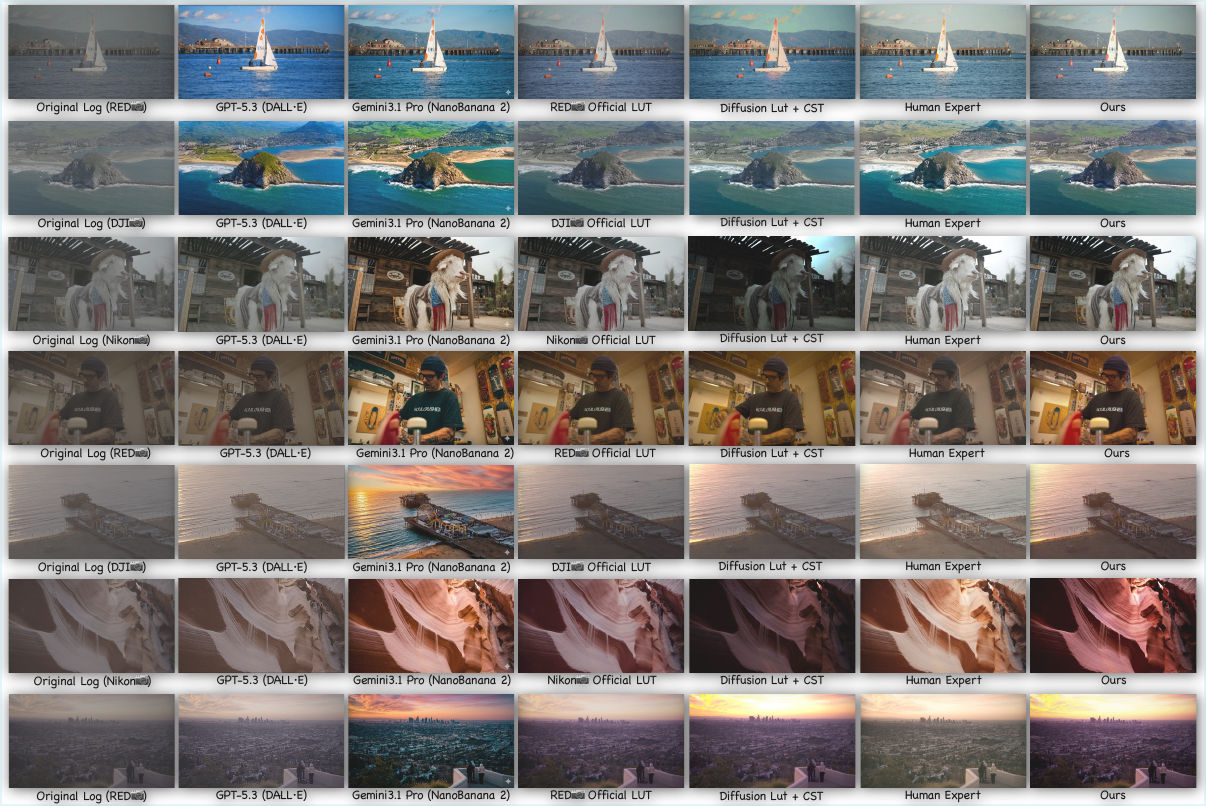}
    \caption{Qualitative comparison with other SOTA methods.}
    \label{fig:convergence}
\end{figure*}

\subsection{Parameter-Driven Execution ($\mathcal{E}$)}
\label{subsec:execution}

The Execution module compiles the Reasoning agent's decisions into a physically realizable output. Rather than generating pixels---as in diffusion-based approaches---the agent translates the deduced ASC-CDL parameters $\mathbf{P}^*$ into a deterministic 3D LUT $\mathcal{L}$.

\textbf{LUT Compilation:} The primary color wheels (Lift $\mathbf{l}$, Gamma $\boldsymbol{\gamma}$, Gain $\mathbf{g}$) are applied sequentially to each lattice node $x_{in} \in [0,1]^3$. We introduce two modifications to the standard ASC-CDL v1.2 pipeline to better suit wide-gamut log workflows: (1)~an \textit{adaptive lift formulation} $x_{lift} = x_{gain} + \mathbf{l} \odot (1.0 - x_{gain})$, where the offset effect smoothly decays as luminance approaches 1.0, preventing highlight clipping in HDR content; and (2)~an \textit{exponential highlight roll-off} for values above a threshold $\tau=0.8$, emulating the shoulder response of analog film to preserve texture in bright regions. The resulting continuous mapping is uniformly sampled into a $33 \times 33 \times 33$ lattice.

\textbf{Agent-Controlled Rendering:} The Execution module operates as the agent's \textit{action interface} to the professional pipeline. After compilation, the agent invokes a virtual rendering environment to apply $\mathcal{L}$ to the anchor frame via hardware-accelerated trilinear interpolation, producing a preview that is fed back to the VLM critic during the ToT evaluation loop (Sec.~\ref{subsec:reasoning}). Upon convergence, the final $\mathcal{L}$ is exported as a standard \texttt{.cube} file, directly importable into professional NLEs such as DaVinci Resolve. The stylized video is rendered as $V_{out} = \mathcal{L}(V_{log})$.

\subsection{State-Aware Reflective Iteration ($\mathcal{F}$)}
\label{subsec:reflection}

While the preceding stages produce a complete cinematic base grade without user intervention, professional color grading is an inherently iterative dialogue between the colorist and the director. The Reflection stage $\mathcal{F}$ enables this dialogue by modeling a parameterized state-transition system. Given a natural language feedback $I_{feedback}$, the agent updates: $\mathbf{P}_{t+1} = \mathcal{F}(\mathbf{P}_t, S, I_{feedback})$.

When feedback is received (e.g., \textit{``Make the shadows slightly cooler''}), the LLM parses the text to explicitly identify the target parameter (in this case, Lift). Crucially, the mapping from linguistic intensity to numerical magnitude is not governed by a hardcoded, rule-based lexicon. Instead, it emerges directly from the LLM's in-context reasoning capabilities. When prompted with a modifier like ``slightly,'' the LLM autonomously deduces a minimal step size (e.g., $\Delta \mathbf{l} \approx \pm 0.01$), whereas ``heavily'' triggers a larger deduced stride (e.g., $\Delta \mathbf{l} \approx \pm 0.05$). To prevent divergence, unmentioned parameters (e.g., Gain) are subjected to a hard constraint and carried over identically from $\mathbf{P}_t$ to $\mathbf{P}_{t+1}$. This structural locking limits the degrees of freedom during iterations, promoting stable convergence. The iterative process terminates either upon user approval or after a predefined maximum number of iterations ($N_{max} = 5$).

\section{The LumiGrade Benchmark}
\label{sec:benchmark}

A critical bottleneck for advancing automated video color grading research is the absence of a suitable public benchmark. Existing datasets used in related work, such as the Condensed Movie Dataset~\cite{bain2020condensed}, consist of already post-produced sRGB clips extracted from released films. While valuable for reference-based style transfer, these datasets bypass the fundamental challenge that professional colorists face daily: interpreting and grading flat, desaturated, log-encoded raw footage directly from the camera sensor. To our knowledge, no public benchmark provides authentic log-encoded video with the accompanying metadata necessary to evaluate end-to-end grading pipelines. To fill this gap, we introduce \textbf{LumiGrade}, a curated evaluation benchmark specifically designed for the task of automated video color grading from raw camera footage.

\textbf{Data Collection.}
We source our footage from professionally shot sample reels released by Mediastorm\footnote{\url{https://www.ysjf.com/material}}. The benchmark comprises over \textbf{100 video clips} spanning a diverse range of cameras, scenes, and shooting conditions. Crucially, all clips are retained in their original camera-native log encodings (e.g., Sony S-Log3, RED Log3G10, ARRI LogC, Panasonic V-Log), preserving the full dynamic range and the characteristic flat, low-contrast appearance that defines the professional grading starting point. Each clip is accompanied by a structured metadata record including: camera model, lens type, log encoding standard, shooting location, time of day, and lighting condition (e.g., natural daylight, golden hour, overcast, interior artificial).

\textbf{Expert Reference Grades.}
To enable quantitative evaluation with ground-truth references, we construct a \textit{reference-graded subset} of 40 representative clips. For each clip in this subset, an experienced colorist performs a full look development pass in DaVinci Resolve, following standard industry practice: primary correction via Lift/Gamma/Gain wheels, followed by creative grading to achieve a polished cinematic look. The final grading state is exported in two forms: (1) a set of ASC-CDL parameters recording the exact mathematical adjustments, and (2) a rendered Rec.709 output video serving as the pixel-level reference. This dual export enables both parameter-space evaluation (comparing predicted ASC-CDL values against expert values) and pixel-space evaluation (computing standard image quality metrics against the rendered reference).

\begin{table*}[t]
\centering
\small
\caption{Quantitative comparison on the LumiGrade benchmark. $\uparrow$: higher is better. Best method in \textbf{bold}, second best \underline{underlined}.}
\label{tab:quantitative}
\begin{tabular*}{\textwidth}{@{\extracolsep{\fill}} l cc cc cccc cc @{}}
\toprule
& \multicolumn{2}{c}{\textbf{Technical Quality}} 
& \multicolumn{2}{c}{\textbf{Aesthetic Quality}} 
& \multicolumn{4}{c}{\textbf{Grading-Specific (LLM-Judge)}} 
& \multicolumn{2}{c}{\textbf{Human Eval}} \\
\cmidrule(lr){2-3} \cmidrule(lr){4-5} \cmidrule(lr){6-9} \cmidrule(l){10-11}
\textbf{Method} 
& MANIQA $\uparrow$ & DeQA $\uparrow$ 
& Q-Align $\uparrow$ & CLIP-AS $\uparrow$ 
& Harmony $\uparrow$ & Tonal $\uparrow$ & Skin $\uparrow$ & Avg. $\uparrow$ 
& Win\% $\uparrow$ & Pro Usab. $\uparrow$ \\
\midrule
GPT-5.4 (DALL·E)             & 0.441 & 2.84 & 3.21 & 5.52 & 4.6 & 4.2 & 4.1 & 4.30 & 1.8  & 1.3 \\
Gemini 3.1 (NanoBanana 2)    & 0.418 & 2.67 & 3.07 & 5.38 & 4.3 & 3.9 & 4.0 & 4.07 & 1.2  & 1.2 \\
Camera Official LUT           & \underline{0.607} & \underline{3.72} & \underline{3.64} & 6.01 & \underline{5.9} & \underline{6.0} & \underline{5.4} & \underline{5.77} & \underline{9.4}  & \underline{3.2} \\
Diffusion Lut + CST           & 0.553 & 3.41 & 3.52 & \underline{6.08} & 5.5 & 5.6 & 5.1 & 5.40 & 5.8  & 2.6 \\
\midrule
Human Expert$^\dagger$         & 0.672 & 4.15 & 4.45 & 6.68 & 7.9 & 7.6 & 7.9 & 7.80 & 43.6 & 4.7 \\
\midrule
\textbf{LumiVideo (Ours)}     & \textbf{0.681} & \textbf{4.22} & \textbf{4.38} & \textbf{6.79} & \textbf{8.1} & \textbf{7.8} & \textbf{7.8} & \textbf{7.90} & \textbf{38.2} & \textbf{4.4} \\
\bottomrule
\end{tabular*}
\vspace{2pt}
\raggedright\footnotesize{$^\dagger$ Non-automated professional reference; excluded from bold/underline ranking.}
\end{table*}

\begin{figure*}[t]
    \centering
    \includegraphics[width=\textwidth]{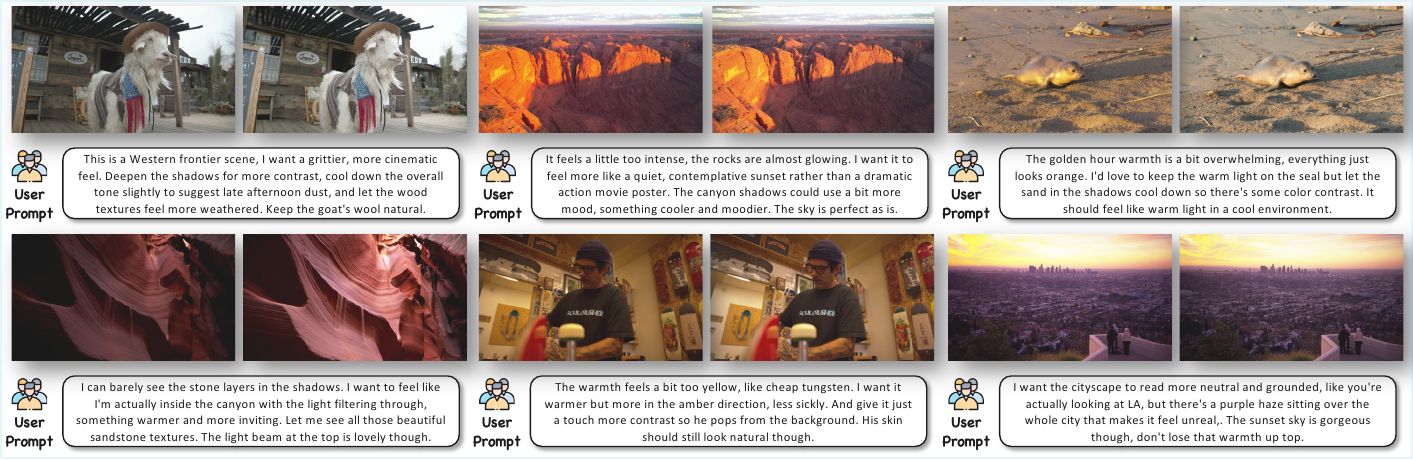}
    \caption{Demonstration of reflective iteration, the automatic base grade is refined via a single natural language directive.}
    \label{fig:convergence2}
\end{figure*}

\section{Experiments}
\label{sec:experiments}

\subsection{Experimental Setup}
\label{subsec:setup}

\textbf{Implementation Details.}
In the \textit{cloud configuration}, the LLM/VLM reasoning stages (Perception, Reasoning, Reflection) are served via API calls to GPT-5.4 and Gemini 3.1 Pro, requiring no local GPU for inference. In the \textit{local configuration}, these modules can be replaced with locally deployed open-source models (e.g., Qwen3-VL) on a single NVIDIA A100 GPU. The RAG vector database is built using OpenAI \texttt{text-embedding-3} and indexed with FAISS~\cite{johnson2019billion}, containing professional grading heuristics curated from cinematography references. For the Tree of Thoughts, we set the maximum depth $D_{max}=2$, branching factor $b=3$, and retain the top-$k=2$ nodes at each pruning step. All 3D LUTs are compiled at $33^3$ resolution with the highlight roll-off threshold $\tau=0.8$. We conducted all experiments using DaVinci Resolve 20\footnote{\href{https://www.blackmagicdesign.com/products/davinciresolve}{https://www.blackmagicdesign.com/products/davinciresolve}}.

\textbf{Baselines.}
We construct a comprehensive set of baselines spanning three categories:

\textit{General-purpose multi-modal models:} We evaluate GPT-5.3 (DALL ·E) and Gemini 3.1 Pro (NanoBanana 2) by providing each model with the same log-encoded anchor frame and a text prompt requesting the target cinematic look. As these models cannot process video, they operate on a per-frame basis.

\textit{Industry baselines:} We include the camera manufacturer's official LUT (e.g., RED IPP2 default) as a representative generic industry transform, and a \textit{Human Expert} grade produced by an experienced colorist in DaVinci Resolve as the professional upper bound.

\textit{Learning-based color grading:} We compare against Shin et al.~\cite{shin2025video}, the SOTA diffusion \cite{ho2020denoising} based method for video color grading. To ensure a strictly fair and rigorous comparison, the reference frames provided to their model are explicitly calibrated to match our foundational Rec.709 Color Space Transform (CST) standard, which named \textit{Diffusion Lut + CST}, thereby eliminating any algorithmic bias caused by color space mismatches.

\textbf{Evaluation Metrics.}
Color grading is inherently a one-to-many task, multiple valid grades exist for any given footage. Consequently, pixel-level fidelity metrics against a single expert reference are ill-suited as primary evaluation criteria. Instead, we evaluate all methods using the non-reference metrics and extensive user studies, organized into three complementary dimensions:

\textit{Technical quality:} MANIQA~\cite{yang2022maniqa} and DeQA \cite{you2025teaching}, which assess low-level perceptual quality such as sharpness, noise, and tonal integrity.

\textit{Aesthetic quality:} Q-Align~\cite{wu2023q} and CLIP Aesthetic Score (CLIP-AS)~\cite{schuhmann2022laion}, which evaluate high-level visual appeal and alignment with human aesthetic preferences.

\textit{Grading-specific assessment:} We employ an LLM-as-Judge protocol using Claude Sonnet 4.6, which scores each result on a 1--10 scale across three grading-specific criteria: \textit{Color Harmony}, \textit{Tonal Range}, and \textit{Skin Tone Naturalness}. The three sub-scores are averaged into a single \textit{LLM-Judge} score. To avoid self-evaluation bias~\cite{zheng2023judging}, we deliberately select this model family entirely independent from our pipeline models.

\subsection{Qualitative Comparison}
\label{subsec:qualitative}

\textbf{Multi-Method Comparison.}
Figure~\ref{fig:convergence} presents qualitative grading results across seven diverse scenes spanning multiple camera formats (RED, DJI, Nikon). All methods receive the same log-encoded input; LumiVideo operates in fully automatic mode without any user directive. Across all scenes, GPT-5.3 (DALL·E) consistently introduces visible spatial artifacts and color hallucinations---most notably altering scene geometry and producing unnatural saturation, confirming that generative pixel-level models are structurally unsuited for color grading. Gemini 3.1 Pro tends to misinterpret the flat log color space, yielding over-contrasty results with crushed shadows and shifted hues. The camera-native Official LUTs (RED, DJI, Nikon) apply technically correct but scene-agnostic transforms that lack creative intent. For example, the DJI LUT renders the coastal scene with a flat, clinical tone that fails to convey the warmth of the natural light. Diffusion LUT + CST produces reasonable results on some scenes but struggles with log-encoded input even after CST pre-processing, exhibiting color casts and inconsistent tonal rendering across different camera formats. The Human Expert achieves the most refined results with nuanced tonal separation tailored to each scene. Our automatic base grade consistently approaches the expert quality: it preserves highlight detail in bright skies, maintains natural skin and fur tones, and autonomously adapts its grading strategy to the detected scene content, applying warm palettes for golden-hour footage, balanced neutral tones for indoor scenes, and appropriate depth separation for landscapes.

\textbf{Reflective Iteration.}
Figure~\ref{fig:convergence2} demonstrates the state-aware refinement loop across six scenes. Starting from the automatic base grade, a single natural language directive is sufficient to steer the result toward the user's specific creative intent. For instance, in the Western frontier scene, the feedback \textit{``I want a grittier, more cinematic feel, keeping the goat's wool natural''} deepens the shadows and shifts the tone cooler while preserving the animal's natural texture. In the canyon interior, the user requests more visible sandstone layers, and the system lifts the shadows to reveal texture without affecting the highlight beam. These examples illustrate that the Reflection loop modifies only the targeted parameters while locking all others, producing predictable, incremental adjustments that converge toward the director's vision.

\begin{table*}[t]
\centering
\small 
\setlength{\tabcolsep}{4pt} 
\caption{Ablation study on the LumiGrade benchmark. All metrics are no-reference. $\uparrow$: higher is better.}
\label{tab:ablation}
\begin{tabular}{l cc cc cccc}
\toprule
& \multicolumn{2}{c}{\textbf{Technical Quality}} & \multicolumn{2}{c}{\textbf{Aesthetic Quality}} & \multicolumn{4}{c}{\textbf{Grading-Specific (LLM-Judge)}} \\
\cmidrule(lr){2-3} \cmidrule(lr){4-5} \cmidrule(lr){6-9}
\textbf{Configuration} & MANIQA $\uparrow$ & DeQA $\uparrow$ & Q-Align $\uparrow$ & CLIP-AS $\uparrow$ & Harmony $\uparrow$ & Tonal $\uparrow$ & Skin $\uparrow$ & Avg. $\uparrow$ \\
\midrule
Full LumiVideo                    & \textbf{0.681} & \textbf{4.22} & \textbf{4.38} & \textbf{6.79} & \textbf{8.1} & \textbf{7.8} & \textbf{7.8} & \textbf{7.90} \\
\midrule
\quad w/o ToT (CoT only)          & 0.634 & 3.52 & 3.91 & 6.34 & 6.8 & 6.5 & 6.5 & 6.60 \\
\quad w/o RAG                     & 0.651 & 3.81 & 4.05 & 6.43 & 7.2 & 6.8 & 6.8 & 6.93 \\
\quad w/o Protected Tones         & 0.654 & 4.06 & 4.12 & 6.51 & 7.6 & 7.3 & 6.2 & 7.03 \\
\quad w/o Adaptive Lift           & 0.632 & 3.68 & 4.21 & 6.58 & 7.7 & 7.4 & 7.2 & 7.43 \\
\quad w/o Reflection (single-shot)& 0.661 & 3.94 & 4.14 & 6.47 & 7.5 & 7.2 & 6.9 & 7.20 \\
\bottomrule
\end{tabular}
\end{table*}

\subsection{Quantitative Evaluation}
\label{subsec:quantitative}
Table~\ref{tab:quantitative} reports the comprehensive evaluation on the full LumiGrade benchmark. LumiVideo achieves the best scores across all metrics by a substantial margin. It further surpasses the Human Expert on the majority of metrics, both technical quality measures, CLIP-AS, and the Harmony and Tonal dimensions of the LLM-Judge, owing to its artifact-free LUT operations and RAG-grounded professional heuristics. The Human Expert retains an edge on Q-Align, Skin Tone Naturalness, and both human evaluation dimensions, reflecting the advantage of manual per-subject refinement for subjective aesthetics and skin-critical scenes. General-purpose generative models lag significantly, confirming that pixel-generation approaches are structurally unsuited for color grading. The Camera Official LUT scores reasonably on technical quality but lower on grading-specific metrics due to its scene-agnostic nature.

\subsection{User Study and Industrial Evaluation}
\label{subsec:userstudy}
To evaluate perceptual quality and industrial applicability, we conduct a two-tier human evaluation study.

\textbf{General User Preference Study.}
We recruit 48 participants with varying levels of photography and video editing experience, ranging from casual mobile editors to semi-professional content creators. Each participant is shown 20 sets of grading results, where each set contains the outputs of all compared methods (presented in randomized order without method labels) alongside the original log-encoded footage. Participants rank the results based on two criteria: (1)~overall aesthetic quality and (2)~faithfulness to the scene intent. As reported in Table~\ref{tab:quantitative}, LumiVideo achieves a win rate of 38.2\%, closely trailing the Human Expert (43.6\%) and far exceeding all other automated baselines. The two generative models collectively account for less than 3.1\% of first-place votes, confirming that pixel-generation approaches are perceptually inadequate for color grading. Notably, the Camera Official LUT and Diffusion Lut + CST receive moderate rankings but are frequently criticized by participants for producing flat or inconsistent tonal rendering across different scene types.

\textbf{Professional Colorist Evaluation.}
We invite 5 professional colorists, each with over 3 years of production experience in commercial and narrative projects, to evaluate a held-out subset of 15 clips. Beyond ranking, colorists rate the usability of each output on a 5-point Likert scale (\textit{``Would you use this as a starting point for further refinement?''}). LumiVideo receives an average usability score of 4.4, approaching the Human Expert (4.7) and substantially above all automated baselines ($\leq$3.2). To further assess real-world integration, we demonstrate the end-to-end workflow: the exported \texttt{.cube} LUT file is imported into DaVinci Resolve's color page and applied to the full timeline. Colorists confirm that the result is immediately editable with standard NLE tools such as Lift/Gamma/Gain wheels and secondary qualifiers---an interoperability that none of the other automated methods provide. In open-ended feedback, colorists consistently highlight two strengths: the system's ability to autonomously produce a usable base grade that respects highlight integrity and skin tones, and the predictability of the natural language refinement loop, which they describe as behaving like a junior colorist who follows directions reliably. The primary limitation noted is the lack of spatially-varying adjustments (e.g., power windows), which we discuss further in Sec.~\ref{sec:conclusion}.

\subsection{Ablation Study}
\label{subsec:ablation}

We conduct ablation experiments on the LumiGrade benchmark to validate the contribution of each key component. Results are summarized in Table~\ref{tab:ablation}.

\textbf{Effect of Tree of Thoughts.}
Replacing the ToT search ($b{=}3$, $D_{max}{=}2$) with a single-pass Chain-of-Thought baseline produces the largest degradation across all metrics, with LLM-Judge dropping from 7.86 to 6.58 ($-$1.28) and Q-Align from 4.38 to 3.91. This confirms that the structured exploration of multiple grading hypotheses is critical for navigating the non-linear ASC-CDL parameter space: a single reasoning chain frequently converges to suboptimal local decisions, particularly on challenging scenes with mixed lighting or high dynamic range.

\textbf{Effect of RAG.}
Removing the RAG module forces the LLM to rely solely on its internalized knowledge, resulting in a moderate but consistent drop (Q-Align $-$0.33, LLM-Judge $-$0.94). Without domain-specific heuristic anchors such as ``teal-orange separation logic'' or ``golden hour warmth recipes'', the model produces less cinematically informed parameter choices---technically adequate but aesthetically less compelling.

\textbf{Effect of Protected Tones.}
Disabling the protected tone constraints has a targeted impact: MANIQA drops from 0.671 to 0.649 and LLM-Judge from 7.86 to 7.03, while MUSIQ remains relatively stable (73.58). This pattern is expected---the global tonal structure is largely preserved, but the LLM-Judge penalizes the contamination of critical hues (particularly skin tones and natural sky gradients), validating that the semantic protection mechanism serves a focused but important role.

\textbf{Effect of Adaptive Lift.}
Reverting to the standard ASC-CDL flat offset causes notable drops in technical quality (MUSIQ $-$2.44, MANIQA $-$0.044), as highlight clipping degrades perceptual sharpness in scenes with bright skies, clouds, or specular reflections. The aesthetic and grading-specific metrics are less affected (LLM-Judge $-$0.41), confirming that the adaptive lift formulation primarily addresses a technical rendering issue rather than a creative one.

\textbf{Effect of Reflection.}
Comparing the single-shot output ($t{=}0$) against the iteratively refined result shows consistent improvement across all metrics, with LLM-Judge increasing from 7.21 to 7.86 ($+$0.65). Figure~\ref{fig:convergence} plots Q-Align scores across reflection iterations, showing a monotonic increase that plateaus around $t{=}2$--$3$, confirming that the state-aware refinement loop converges efficiently within a small number of iterations.

\section{Conclusion}
\label{sec:conclusion}

We presented LumiVideo, an agentic system that reformulates video color grading as a transparent, parameter-driven reasoning process. Given only raw log-encoded footage, LumiVideo autonomously produces a cinematic base grade through four stages---Perception, Reasoning, Execution, and Reflection---outputting industry-standard ASC-CDL parameters and \texttt{.cube} LUT files that integrate directly into professional workflows. An optional natural language Reflection loop enables precise, incremental refinement. We further introduced LumiGrade, the first publicly available log-encoded video benchmark for evaluating automated grading. Experiments demonstrate that LumiVideo surpasses human expert quality on the majority of metrics while enabling iterative creative control.

\textbf{Limitations.}
Our current system applies a single global LUT to the entire frame, which cannot address scene-local adjustments such as per-subject color isolation, the operations that professional colorists achieve through Power Windows and secondary qualifiers. Extending the action space to include spatially-varying parameters is a promising direction, would further benefit the community.

\section*{Acknowledgments}
This work was supported in part by the Guangdong Provincial Key Laboratory of IRADS (2022B1212010006), in part by the Guangdong Higher Education Upgrading Plan (2021–2025), in part by the Guangdong and Hong Kong Universities “1+1+1” Joint Research Collaboration Scheme, in part by the National Key Research and Development Program of China (2022ZD0117700), in part by the National Natural Science Foundation of China (62325204), and in part by the MSAI program of Northwestern University. The authors would like to thank Mediastorm's free video samples.

\bibliographystyle{ACM-Reference-Format}
\bibliography{sample-base}

\end{document}